\documentclass[11pt,letterpaper]{article}
\usepackage{emnlp2016,graphicx}
\usepackage{times,algorithmic,algorithm}
\usepackage{latexsym,subfig}
\usepackage{amsmath,bm,amssymb}

\usepackage{times}
\usepackage{latexsym}

\emnlpfinalcopy

\def\emnlppaperid{***}

\newcommand{\mat}[1]{\mathbf{#1}}

\newcommand{\vsuper}[2]{\mathbf{#1}^{(\text{#2})}}

\newcommand{\msub}[2]{\mat{#1}_{\text{#2}}}
\newcommand{\vsupersub}[3]{\mathbf{#1}^{(\text{#2})}_{\text{#3}}}

\newcommand\mef{\vsuper{m}{k}}

\newcommand\mP{\mat{P}}

\newcommand\sent{\mat{c}}
\newcommand\Corp{\mat{C}}

\newcommand\softmax{\operatorname{softmax}}

\newcommand\Wr{\msub{W}{r}}

\newcommand\Ur{\msub{U}{r}}

\newcommand\Vr{\msub{V}{r}}
\newcommand\Wi{\msub{W}{i}}
\newcommand\Ui{\msub{U}{i}}

\newcommand\Vg{\msub{V}{g}}
\newcommand\Ug{\msub{U}{g}}

\newcommand\xstu{\vsupersub{x}{k}{m}}

\newcommand\vs{\mathbf{b}^{T}}
\newcommand\vns{\mathbf{b}}

\newcommand\hst{\vsupersub{u}{k}{m}}

\newcommand\hsT{\vsuper{u}{k}}

\newcommand\hik{\vsuper{z}{k}}
\newcommand\hikm{\vsuper{z}{k-1}}

\newcommand\htt{\vsupersub{v}{k}{n}}

\newcommand\httm{\vsupersub{v}{k}{n-1}}

\newcommand\yt{\vsuper{y}{k}}
\newcommand\ytt{\vsupersub{y}{k}{n}}

\newcommand\yttm{\vsupersub{y}{k}{n-1}}

\newcommand\ctt{\vsupersub{a}{k}{n}}

\newcommand\rtt{\vsupersub{r}{k}{n}}

\newcommand\rttm{\vsupersub{r}{k}{n-1}}
\newcommand\vxs{\vsuper{x}{k}}

\newcommand\vxsu{\vsuper{x}{k}}

\newcommand\vwtu{\vsuper{w}{k}}

\newcommand\Ua{\msub{U}{a}}

\newcommand\Va{\msub{V}{a}}

\title{An Attentional Neural Conversation Model with Improved Specificity}

\author{Kaisheng Yao\\
	    Microsoft Research\\
	    Redmond, USA \\
	    {\footnotesize{\tt kaisheny@microsoft.com}}
  \And
  Baolin Peng \\
  Chinese University \\ of Hong Kong\\
  {\footnotesize{\tt blpeng@se.cuhk.edu.hk}}
	  \And
	Geoffrey Zweig\\
  	Microsoft Research\\
  	Redmond, USA\\
  {\footnotesize{\tt gzweig@microsoft.com}}
	  \And
	Kam-Fai Wong\\
  	Chinese University \\ of Hong Kong\\
  	{\footnotesize{\tt kfwong@se.cuhk.edu.hk}}
}
\date{}

\begin{document}

\maketitle

\begin{abstract}
In this paper we propose a neural conversation model for conducting dialogues. We demonstrate the use of this model to generate help desk responses, where users are asking questions about PC applications. Our model is distinguished by two characteristics. First, it models intention across turns with a recurrent network, and incorporates an attention model that is conditioned on the representation of intention.  Secondly, it avoids generating non-specific responses by incorporating an IDF term in the objective function. The model is evaluated both as a pure generation model in which a help-desk response is generated from scratch, and as a retrieval model with performance measured using recall rates of the correct response. Experimental results indicate that the model outperforms previously proposed neural conversation architectures, and that using specificity in the objective function significantly improves performances for both generation and retrieval.

\end{abstract}

\section{Introduction}
In recent years, neural network based conversation models~\cite{Serban2015Dialogue,Sordoni2015HRED,Vinyals2015NeuralConversationModel,Shang2015NRM} have emerged as a promising complement to traditional partially observable Markov decision process (POMDP) models~\cite{Young2013POMDP}. The neural net based techniques require little in the way of explicit linguistic knowledge, e.g., creating a semantic parser, and therefore have promise of scalability, flexibility, and language independence.
Broadly speaking, there are two approaches to building a neural conversation model. The first is to train what is essentially a conversation-conditioned language model, which is used in generative mode to produce a likely response to a given conversation context. The second approach is retrieval-based, and aims at selecting a good response from a list of candidates taken from the training corpus.

Neural conversation models are usually trained similarly to neural machine translation models~\cite{sutskever2014sequence,Cho2014Seq2Seq}, which treat response generation as a surface-to-surface transformation. While simple, it is possible that these models would benefit from explicit modeling of conversational dynamics, specifically the attention and intention processes hypothesized in discourse theory~\cite{Grosz1986Discourse}. A recent improvement along these lines is the hierarchical recurrent encoder-decoder in~\cite{Serban2015Dialogue,Sordoni2015HRED}  that incorporates two levels of recurrent networks, one for generating words and one for modeling dependence between conversation turns. In this paper, we extend this approach further, and propose an explicit intention/attention model.

We further tackle a second key problem of neural conversation models, their tendency to generate generic, non-specific responses~\cite{Vinyals2015NeuralConversationModel,Li2016MMI}. To address the problem of specificity, a maximum mutual information (MMI) method for generation was proposed in ~\cite{Li2016MMI}, which models {\it both} sides of the conversation, each conditioned on the other. While we find this method effective, training the additional model doubles the computational cost.

The contributions of this paper are as follows. First, we introduce a novel attention with intention neural conversation model that integrates an attention mechanism into a hierarchical model. Without the attention mechanism, the model resembles the models in \cite{Serban2015Dialogue} but performs better. Visualization of the intention layer in the model shows that it indeed is relevant to intent.
Second, we address the specificity problem by incorporating inverse document frequency (IDF)~\cite{Salton1988TFIDF,Ramos2003TFIDF} into the training process. The proposed training algorithm uses reinforcement learning with the IDF value of the generated sentences as a reward signal. To the best of our knowledge, this is the first method incorporating specificity into the training objective function. Empirically, we find that it performs better than the dual-model method of~\cite{Li2016MMI}. Lastly, we demonstrate that the proposed model also performs well for retrieval-based conversation modeling. Using a recently proposed evaluation metric~\cite{Lowe2015Ubuntu,Lowe2016NUC}, we observed that this model was able to incorporate term-frequency inverse document frequency (TF-IDF)~\cite{Salton1988TFIDF} and significantly outperformed a TF-IDF retrieval baseline and the model without using TF-IDF.

\section{The model}
\label{sec:awimodel}

\begin{figure}[t]
\centering
\includegraphics[width=0.92\columnwidth]{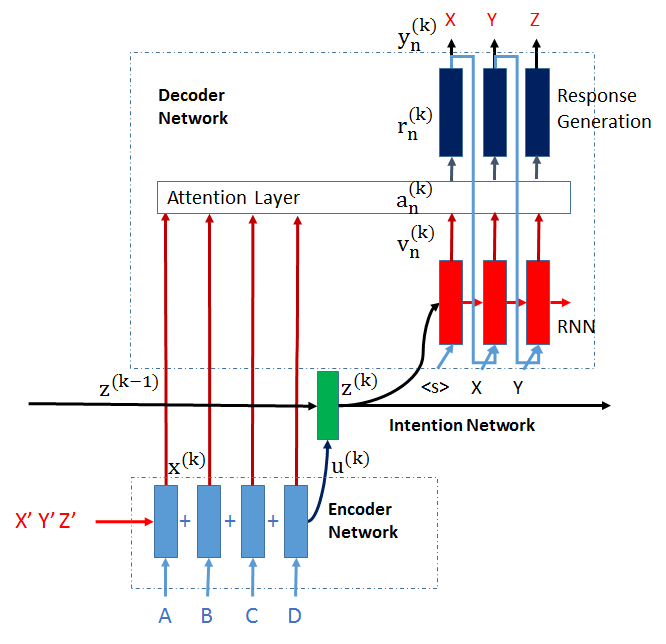}
\caption{Illustration of the AWI model in one turn to generate a response sequence [X,Y,Z] for input sequence [A,B,C,D] and the previous response [X',Y',Z'].
}
\label{fig:attention}
\vspace{-10pt}
\end{figure}

The proposed model is in the encoder-decoder framework~\cite{sutskever2014sequence} but incorporates a hierarchical structure to model dependence between turns of conversation process. The encoder network processes the user input and represents it as a vector. This vector is the input to a recurrent network that models context or intention to generate response in the decoder network. The decoder generates a response sequence word-by-word. For each word, the decoder uses an attention model on the words in the user input. 
Following \cite{Grosz1986Discourse}, we refer the conversation context as intention. Because an attention model is used in the decoder, we denote this model as attention with intention (AWI) neural conversation model.
A detailed illustration for a particular turn is in Figure~\ref{fig:attention}.
We elaborate each component of this model in the following sections. 

\subsection{Encoder network}
\label{sec:encoder}
Given a user input sequence with length $M$ at turn $k$, the encoder network converts it into a sequence of vectors $\vxsu = [\xstu : m = 1, \cdots, M]$, with vector $\xstu \in R^{d_e}$ denoting a word embedding representation of the word at position $m$. 
The model uses a feed-forward network to process this sequence. It has two inputs. The first is a simple word unigram feature, extracted as the average of the word embeddings in $\vxsu$. The second input is a representation of the previous response. This representation is also a word unigram feature, but is applied on the past response. The output, $\hsT \in R^{d_x}$, from the top layer of the feed-forward network is a vector representation of the user input. 
In addition to this vector representation, the encoder network outputs the vector sequence $\vxsu$. 

\subsection{Intention network}
\label{sec:intention}
The middle layer represents a conversation context, which we denote it as the intention hypothesized in~\cite{Grosz1986Discourse}. At turn $k$, it is a vector $\hik \in R^{d_z}$. 
To model turn-level dynamics, the activity in $\hik$ is dependent on the activity in the previous turn $\hikm$ and the user input at the current turn. Therefore, it is natural to represent $\hik$ as follows
\begin{equation}
\hik = \tanh\left(\Wi \hikm + \Ui \hsT\right),
\end{equation}
\noindent where $\tanh()$ is a tanh operation. $\Wi \in R^{d_z \times d_z}$ and $\Ui \in R^{d_z \times d_x}$ are matrices to transform inputs to a space with dimension $d_z$. Usually, we apply multiple layers of the above non-linear processing, and the higher layer only applies $\Wi$ to the output from the layer below. Notice that $\Wi$s are untied across layers. The output from the top one is the representation of the intention network.

\subsection{Decoder network}
\label{sec:decoder}
Decoder network has output $\ytt \in R^{V }$, a vector with dimension of vocabulary size $V$. Each element of the vector represents a probability of generating a particular word at position $n$. This probability is conditioned on $\yttm$, the word generated before, the above described intention vector $\hik$ and the encoder output $\vxs$. To compute this probability, it uses softmax on a response vector $\rtt \in R^{V}$. The probability is defined as follows
\begin{eqnarray}
p(\ytt|\yttm,\hik,\vxs) & =  & \softmax(\rtt). \label{eqn:prob}
\end{eqnarray}
The decoder uses the following modules to generate response vector $\rtt$. 

\paragraph{RNN}
\label{sec:decoderhiddenstate}
For position $n$, the hidden state $\htt$ of the RNN is recursively computed as 
\begin{equation}
\htt = \tanh \left( \Vr \httm + \Wr \yttm + \Ur \rttm \right), \label{eqn:rnn}
\end{equation}
\noindent where $\httm$, $\yttm$ and $\rttm$ each represent the previous hidden state, the word generated and the response vector for word position $n-1$. $\Vr \in R^{d_r \times d_r}$, $\Wr \in R^{d_r \times d_V}$ and $\Ur \in R^{d_r \times d_V}$ are matrices to transform their right side inputs to a space with dimension $d_r$. During training, $\yttm$ is a one-hot vector with its non-zero element corresponding to the index of the word at $n-1$. During test, it is still a one-hot vector but the index of the non-zero element is from beam search or greedy search at position $n-1$. 
We apply multiple layers of the above process on $\htt$, with the higher layer uses only the lower layer output in the left side of (\ref{eqn:rnn}). Parameters such as $\Vr$ are untied across layers. The top level response in (\ref{eqn:rnn}) is the RNN output. 
To incorporate conversation context, the initial state $\htt$ at $n=0$ is set to $\hik$ from the intention network.

\paragraph{Attention layer}
\label{sec:attention}

We use the content-based attention mechanism~\cite{bahdanau2015neural,LuongEMNLP2015}. It is a single layer neural network that aligns the target side hidden state $\httm$ at the previous position $n-1$ with the source side representation $\hst$ at word position $m$. The alignment weight $w_{y_nm}$ is computed as follows
\begin{eqnarray}
e_{y_nm} & = & \vs \tanh(\Va \httm + \Ua \xstu), \\
w_{y_nm} & = & \frac{\exp{e_{y_nm}}}{\sum_{j=1}^M \exp(e_{y_nj})},  \label{eq:alignment2} \\
\ctt & = & \sum_m w_{y_nm} \xstu, \label{eqn:attentionoutput}
\end{eqnarray}
\noindent where $\Va \in R^{d_a \times d_V}$ and $\Ua \in R^{d_a \times d_x}$ are matrices to transform their inputs to a space with dimension $d_a$. $\vns \in R^{d_a}$ is a vector. The softmax operation in (\ref{eq:alignment2}) is normalized on the user input sequence. $\ctt$ in (\ref{eqn:attentionoutput}) is the output of the attention layer.

\paragraph{Response generation}
\label{sec:response}
We use the following feed-forward network to generate a response vector $\rtt \in R^{d_V}$, using the decoder RNN output $\htt$ and the attention layer output $\ctt$; i.e., 
\begin{eqnarray}
\rtt & = & \tanh\left(\Vg \htt + \Ug \ctt\right), \label{eqn:response}
\end{eqnarray}
\noindent where $\Vg \in R^{d_V \times d_r}$ and $\Ug \in R^{d_V \times d_a}$ are matrices to transform their right side inputs to a space with dimension $d_V$. Similarly as those networks described above, we use untied multiple layers to generate $\rtt$, and the input to the higher layer only use the output from the layer below. The top layer output is fed into softmax operation in Eq. (\ref{eqn:prob}) to compute the probability of generating words. 

\paragraph{Input similarity feature}
\label{sec:mefeature}
We construct a linear direct connection between user inputs and the output layer. To achieve this, a large matrix $\mP$ with dimension $V \times V$ is used to project each input word to a high dimension vector with dimension size $V$. The projected words are then averaged to have a vector $\mef$ with dimension $V$. This vector is added to the response vector $\rtt$, so that Eq. (\ref{eqn:prob}) is computed as $\softmax(\rtt + \mef)$. Since $\mef$ is applied to all word positions at turn $k$, it provides a global bias to the output distribution. 

\section{Training and decoding algorithms}
\label{sec:training}
This section presents training and decoding algorithms for response generation and retrieval. Section~\ref{sec:xent} is the standard cross-entropy training. It is used for training both generation and retrieval models. Section \ref{sec:reinforce} introduces training and decoding algorithms to enhance specificity for generation. Algorithms for training and decoding for retrieval are described in Sec.~\ref{sec:trainranking}. 

\subsection{Maximum-likelihood training}
\label{sec:xent}
The standard training method maximizes the probability of predicting the correct word $\ytt$ given user input $\vxs$, context $\hik$, and the past prediction $\yttm$; i.e., the objective is maximizing the following log-likelihood w.r.t. the model parameter $\theta$,
\begin{equation}
L(\yt) = \log \prod_{n=1}^N p(\ytt|\yttm,\hik,\vxs;\theta). \label{eqn:xent}
\end{equation}
 

A problem with this training is that the learned models are not optimized for the final metric~\cite{Och2003MERT,Ranzato2016ICLR}. Another problem is that decoding to maximize sentence likelihood typically results in non-specific high-frequency words~\cite{Li2016MMI}.

\subsection{Improving specificity for generation}
\label{sec:reinforce}
We propose using inverse document frequency (IDF)~\cite{Salton1988TFIDF} to measure specificity of a response. IDF is used in decoding in Sec.~\ref{sec:rerankingwithidf}. We describe a novel algorithm in Sec.~\ref{sec:gradwithrewards} that incorporates IDF in training objective. 
\subsubsection{Specificity} 
\label{sec:specificity}
IDF for a word $w$ is defined as 
\begin{equation}
idf(w) = \log \frac{N}{|{ \sent \in \Corp: w \in \sent}|},
\end{equation}
\noindent where $N$ is the number of sentences in a corpus $\Corp$ and $\sent$ denotes a sentence in that corpus. The denominator represents the number of sentences in which the word $w$ appears. 
A property of IDF is that words occur very frequently have small IDF values. 

We further define a sentence-level IDF as the average of IDF values of words in a sentence; i.e.,
\begin{equation}
idf(\sent) = \frac{1}{|\sent|} \sum_{w \in \sent} idf(w), \label{eqn:sentenceidf}
\end{equation}
\noindent where the denominator $|\sent|$ is the number of occurrence of words in sentence $\sent$. A corpus-level IDF value is similarly computed on a corpus with an average operation as $idf(\Corp) = \frac{1}{|\Corp|} \sum_{\sent \in \Corp, w \in \sent} idf(w)$ and the denominator in the equation is the number of occurrence of words in the corpus. 

\subsubsection{Reranking with IDF}
\label{sec:rerankingwithidf}
One way to improve specificity is using IDF to rerank hypotheses from beam search decoding. 
The length-normalized log-likelihood scores of these hypotheses are interpolated with sentence-level IDF scores. Tuning the interpolation weight is on a development set using minimum error rate training (MERT)~\cite{Och2003MERT}.  
The interpolation weight that achieves the highest BLEU~\cite{Papineni2002BLEU} score on the development set is used for testing.

\subsubsection{Incorporating IDF in training objective}
\label{sec:gradwithrewards}
Alternatively, we cast our problem of optimizing a model directly for specificity in the reinforcement learning framework~\cite{reinforcementlearning}. The decoder is an agent with its policy from Eq. (\ref{eqn:prob}). Its action is to generate a word using the policy, and therefore it has $V$ actions to take at each time. At the end of generating a whole sequence of words for response, the agent receives a reward, calculated as the sentence level IDF score of the generated response. Training therefore should find a policy that maximizes the expected reward. 

This problem can be solved using REINFORCE~\cite{Williams1992REINFORCE}, in which the gradient to update model parameter is calculated as follows
\begin{eqnarray}
\lefteqn{\Delta \theta \propto \left(r(\vwtu) - b(\vxs)\right)} \label{eqn:reinforce}  \\
& & \frac{\partial \log \prod_n p(\ytt|\yttm,\hik,\vxs;\theta))}{\partial \theta}, \nonumber
\end{eqnarray}
\noindent where $r(\vwtu) = idf(\vwtu)$ is the IDF score of the generated response $\vwtu$ at turn $k$. $b(\vxs)$ is called reinforcement baseline. It in practice can be set empirically as an arbitrary number to improve convergence~\cite{Zaremba2015NTM}. 
One convenient way of estimating the baseline is the mean of the IDF values on the training set, which is used in this paper.

Notice that the IDF score is computed on the decoded responses $\vwtu$, but the log-likelihood is computed on the correct response. Therefore, the algorithm improves likelihood of the correct response and also encourages generating responses with high IDF scores.

\subsection{Training and decoding for retrieval}
\label{sec:trainranking}
The conversation model can be used for retrieval of the correct responses from candidates. 
We briefly describe TF-IDF in Sec.~\ref{sec:tfidf}. Section~\ref{sec:rankingcriterion} presents the algorithm to train AWI model for retrieval. Section~\ref{sec:decodingir} combines the model with TF-IDF. 
Notice that TF-IDF uses IDF to penalize non-specific words, combining the AWI model with TF-IDF should have improved specificity, which could lead to improved performances for retrieval.

\subsubsection{TF-IDF}
\label{sec:tfidf}
Term-frequency inverse document frequency (TF-IDF)~\cite{Salton1988TFIDF} is an established baseline for conversation model used for retrieval~\cite{Lowe2015Ubuntu}. The term-frequency (TF) is a count of the number of times a word appears in a given context, and IDF puts penalty on how often the word appears in the corpus. The TF-IDF is a vector for a context computed as follows for a word $w$ in a context $\sent$, 
\begin{equation}
tfidf(w, \sent, \Corp) = o(w,\sent) \times \log \frac{N}{|{\sent \in \Corp: w \in \sent}|},\nonumber
\end{equation}
\noindent where $o(w,\sent)$ is the number of times the word $w$ occurs in the context $\sent$. For retrieval, a vector for a conversation history is firstly created, with element computed as above for each word in the conversation history. Then, a TF-IDF vector is computed for each response. Similarity of these vectors are measured using Cosine similarity. The responses with the top $k$ similarities are selected as the top $k$ outputs. 

\subsubsection{Training models with ranking criterion}
\label{sec:rankingcriterion}
In order to train AWI model for retrieval, the model needs two types of responses. The positive response is the correct one, and the negative responses are those randomly sampled from the training set. 
For a response $\vwtu$, its length-normalized log-likelihood is computed as follows,
\begin{equation}
llk(\vwtu) = \frac{L(\vwtu)}{|\vwtu|},
\end{equation}
\noindent where $L(\vwtu)$ is computed using Eq. (\ref{eqn:xent}) with $\yt$ substituted by $\vwtu$. $|\vwtu|$ is the number of words in $\vwtu$.

The objective is to have high recall rates such that the correct responses are ranked higher than negative responses. To this end, the algorithm maximizes the difference between the length-normalized log-likelihood of the correct responses to the length-normalized log-likelihood of negative responses; i.e.,
\begin{equation}
R = \max \{llk(\yt) - llk(\vwtu) \},
\end{equation}
\noindent where $\yt$ is the correct response and $\vwtu$ is the negative response. 


\subsubsection{Ranking with AWI together with TF-IDF} 
\label{sec:decodingir}
Naturally, the length-normalized log-likelihood score from the model trained in Sec.~\ref{sec:rankingcriterion} can be interpolated with the similarity score from TF-IDF in Sec.~\ref{sec:tfidf}. 
The optimal interpolation weight is selected on a development set if it achieves the best recall rates.

\section{Experiments}
\label{sec:experiments}
\subsection{Data}
\label{exp:data}
We use a real-world commercial data set to evaluate the models. The data contains human-human dialogues from a helpdesk chat service for services of Office and Windows. In this service, costumers seek helps on computer and software related issues from human agents. 
Training set consists of 141,204 dialogues with 2 million turns. The average number of turns in a dialogue is 12, with the largest number of 140 turns and the minimum number of 1 turn. More than 90\% of dialogues have 25 or fewer turns. The number of tokens is 25,410,683 in the source side and is 37,796,000 in the target side. The vocabulary size is 8098 including words from both source and target sides. 
Development and test sets each have 10,000 turns. The test set has 125,451 tokens in its source side and 187,118 in its target side. 


\subsection{Training Details}
\label{exp:details}
Unless otherwise stated, all of the recurrent networks have  two layers. The encoder network uses word embedding initialized from 300-dimension GLOVE vector~\cite{pennington2014glove} trained from 840 billion words. Therefore, the embedding dimension $d_e$ is 300. The hidden layer dimension for encoder, $d_x$, is 1000. Decoder dimension, $d_r$, is 1000. The intention network has a 300 dimension vector; i.e., $d_z = 300$. The alignment dimension $d_a$ is 100. All parameters, except biases, are initialized using uniform distribution in $[-1.0, 1.0]$ but are scaled to be inversely proportional to the number of parameters. All bias vectors are initialized to zero. 

The maximum number of training epochs is 10.  We use RMSProp with Momentum~\cite{tieleman2012rmsprop} to update models. 
We use perplexity to monitor the progress of training; learning rate is halved if the perplexity on the development set is increased. 
The gradient is re-scaled whenever its norm exceeds 5.0. 
To speed-up training, dialogues with the same number of turns are processed in one batch. The batch size is 20. 

Hyper-parameters such as initial learning rate and dimension sizes are optimized on the development set. These parameters are then used on the test set.  Decoding uses beam search with a beam width of 1. Decoding stops when an end of sentence is generated.

\subsection{Evaluation metrics}
\label{exp:metric}
As our model is used both for generation and retrieval, we use some established measures in the literature for comparison. 
The first measure is BLEU~\cite{Papineni2002BLEU}, which uses the N-gram~\cite{Dunning1994ngram} to compute similarities between references and responses, together with a penalty for sentence brevity. We use BLEU with 4-gram. 
While BLEU may unfairly penalize paraphrases with different wording, it has found correlated well with human judgement on responses generation tasks~\cite{galley2015bleu}. The second measure is perplexity~\cite{Brown1992perplexity}, which measures the likelihood of generating a word given observations. We use it in section~\ref{exp:others} to compare the proposed model with respect to other neural network models that also report perplexity. However, since our training algorithms in Sec.~\ref{sec:reinforce} is designed to improve specificity, which is not directly correlated with the standard likelihood, we only report perplexity in section~\ref{exp:others}. 
The third metric is corpus level IDF score for specificity, computed in (\ref{eqn:sentenceidf}). 

Since our model is also used for retrieval, we adopt a response selection measure proposed in \cite{Lowe2015Ubuntu}, in which the performance of a conversation model is measured by the recall rate of those correct responses in the top ranks. This metric is called Recall@k (R@k). The model is asked to select the $k$ most likely responses, and it is correct if the true response is among these $k$ responses. The number of candidates for retrieval is 10, following  \cite{Lowe2015Ubuntu}. 
This measure is observed to correlate well with human judgment for retrieval based conversation model~\cite{Lowe2016NUC,Liu2016Dialogue}. 

\subsection{Performance as a generation model}

\subsubsection{Comparison with other methods}
\label{exp:others}
\begin{table}[t]
\begin{center}
\begin{tabular}{|p{4.0cm}|c|c|}
\hline
Models & BLEU & Perplexity \\
\hline \hline
N-gram & 0.0 & 280.5 \\
\hline
Seq2Seq \cite{Vinyals2015NeuralConversationModel} & 1.82 & 12.64 \\
\hline 
HRED \cite{Serban2015AAAI} & 6.14 & 13.82 \\
\hline 
AWI & 9.29 & 11.52 \\
\hline
\end{tabular}
\end{center}
\caption{Results on the test set.  \vspace{-10pt}
\label{tab:comparison}} 
\end{table}

We compared the AWI model with the sequence-to-sequence (Seq2Seq)~\cite{Vinyals2015NeuralConversationModel} and the hierarchical recurrent encoder-decoder (HRED)~\cite{Serban2015AAAI} models. All of the models had a two layers of encoder and decoder. 
The hidden dimensions for the encoder and decoder were set to 200 in all of the models. The hidden dimension for the intention network was set to 50.  
All of the models had their optimal setup on the development set. Both Seq2Seq and HRED used long short-term memory networks~\cite{Hochreiter97}. The number of parameters was approximately $4.48 \times 10^6$ for Seq2Seq and $4.50 \times 10^6$ for HRED. AWI didn't have the input similarity feature and it had $5.71 \times 10^6$ parameters.
Greedy search was used in this experiment. 

Table~\ref{tab:comparison} shows that AWI model outperforms other neural network models both in BLEU and perplexity. For comparison, BLEU and perplexity scores from an unconditional N-gram model are also reported, which are much worse than those from neural network based models. 
The BLEU score for n-gram was obtained by sampling from the n-gram and comparing the sampled response to a typical response. Its response has BLEU score of 0.08 if using BLEU with 1-gram. Because it cannot be matched in 4 gram with the typical response, its BLEU with 4 gram is 0.  
On one experiment, not shown in the table due to space limitation, with smaller training set, we observed these models performed worse yet similarly. This suggests that the benefit of incorporating hierarchical structure in both HRED and AWI is more apparent with larger training data. 

\subsubsection{Results with specificity improved models}
\label{exp:idf}
\begin{table}[t]
\begin{center}
\begin{tabular}{|c|c|c|}
\hline
Models & BLEU & IDF \\
\hline \hline
AWI  & 11.42 & 2.35 \\
\hline 
AWI + sampling & 7.02 & 2.76 \\
\hline
AWI + MMI \cite{Li2016MMI} & 11.47 & 2.37 \\
\hline
AWI + IDF & 11.43 & 2.83 \\
\hline
IR-AWI & 11.70 & 2.40 \\
\hline
\end{tabular}
\end{center}
\caption{Performance for generation. \label{tab:reinforce}} 
\vspace{-10pt}
\end{table}

We report BLEU and IDF scores in table~\ref{tab:reinforce}. 
The baseline is AWI trained with standard cross-entropy in Sec.~\ref{sec:xent}. 
For comparison, we used a sampling method~\cite{RNN} to generate responses, denoted as "AWI + sampling". Using sampling would lead to an appropriate number of infrequent words and therefore an IDF score that is similar to that of the reference responses. Indeed this is observed in "AWI + sampling" in the table. It has an IDF score of 2.76, close to the IDF score of 2.74 from the training set. 
However, sampling produces worse BLEU scores, though it has higher IDF score than AWI. 

We also report result using MMI method for decoding~\cite{Li2016MMI}, denoted as "AWI + MMI". This uses a backward directional model trained for generating source from target, and its decoding uses reranking algorithm in Sec.~\ref{sec:rerankingwithidf}. The optimal interpolation weight for the backward directional model was 0.05. Both BLEU and IDF scores are improved. 
On the other hand, "AWI + MMI" requires an additional model  as complicated as the baseline AWI model. 

Alternatively, AWI results are reranked with the sentence-level IDF scores using algorithm in Sec.~\ref{sec:rerankingwithidf}. The optimal interpolation weight to IDF was 0.035. This result, denoted as "AWI + IDF", has improved BLEU and IDF scores, in comparison to the baseline AWI. Compared against "AWI + MMI", it has similar BLEU but higher IDF scores. This suggests that IDF score is more directly related to specificity than using MMI. 

The result from using specificity as reward to train a model in Sec.~\ref{sec:gradwithrewards} is denoted as IDF-rewarded AWI model or "IR-AWI".
The reinforcement baseline $b(\vxs)$ was empirically set to 1.0. 
We also experimented with a larger value to 1.5 for the baseline and didn't observe much performance differences. 
"IR-AWI" consistently outperforms "AWI" and "AWI + MMI". 

\subsubsection{Analysis}
\label{exp:analysisawi}

\begin{figure}[t]
\centering
\vspace{-15pt}
\includegraphics[width=0.5\textwidth]{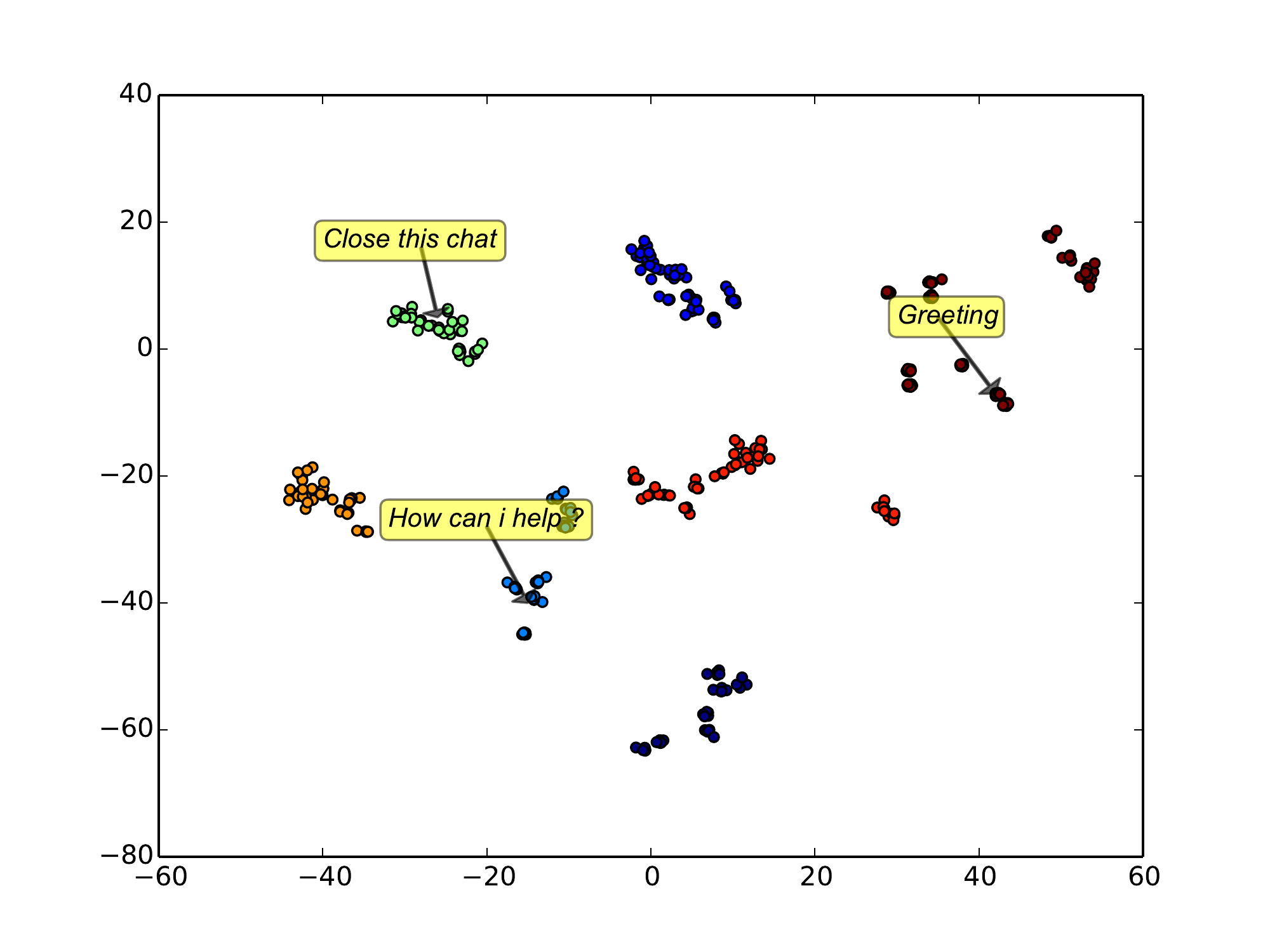}
\caption{t-SNE visualization of intention vectors. }
\label{fig:tsneintention}
\vspace{-10pt}
\end{figure}

Figure~\ref{fig:tsneintention} uses t-SNE~\cite{tSNE} to visualize the intention vectors. It shows clear clusters even though training intention vectors doesn't use explicit labels. In order to relate these clusters with explicit meaning, we look at the responses generated from these intention vectors, and tag similar responses with the same color. Some responses types are clearly clustered such as "Greeting" and "Close this chat". Others types are though more distributed and cannot find a clear tag for these responses. We therefore leave them untagged. 

We also show two examples of responses in Tables~\ref{tab:sampleresponse} and \ref{tab:sampleresponse2} from AWI and IR-AWI. Responses from AWI+MMI and AWI+IDF are the same as from AWI, so we only list responses from AWI and IR-AWI. These responses are reasonable. However, the IR-AWI responses in these tables are more specific than the generic responses from AWI. 

\begin{table}[t]
\begin{center}
\begin{tabular}{|p{3.5cm}|p{3.5cm}|}
\hline
AWI & IR-AWI \\
\hline 
okay, thank you for that information. & alright, kindly click the link below and update me once you are on the page that is asking for a six digit code http://$\langle$webpage$\rangle$. \\ 
\hline 
\end{tabular}
\end{center}
\caption{Examples of responses. Conversation history is as follows. User said, "i don't have another computer that supports miracast. the adapter appears as a 'device' but does not get connected. do you want remote access to it?" Agent replied, "would that be alright for you ?" User then said, "yes."  For reference, the human agent responds, "alright, kindly click the link below and update me once you are on the page http://$\langle$webpage$\rangle$, i would like to set your expection." 
\label{tab:sampleresponse}} 
\end{table}

\begin{table}[t]
\begin{center}
\begin{tabular}{|p{3.5cm}|p{3.5cm}|}
\hline
AWI & IR-AWI \\
\hline 
may i know how did you upgrade to windows 10?& may i have the product key for windows 8.1?\\
\hline 
\end{tabular}
\end{center}
\caption{Examples of responses. Conversation history is as follows. User said, "windows activate error code : $\langle$errorcode$\rangle$," Agent replied, "i am sorry for the inconvenience. but nothing to worry about, i will surely help you with this. may i know the previous os?" User then said, "8.1."  For reference, the human agent responds, "okay, may i know, do you have a product key for windows 8.1?"
\label{tab:sampleresponse2}} 
\end{table}


\subsection{Performance for retrieval}
\label{exp:classification}
We report recall rates, R@1 and R@5, in Table~\ref{tab:ranking}. Clearly, AWI model trained with ranking criterion in Sec.~\ref{sec:rankingcriterion} outperforms TF-IDF. Importantly, AWI model was able to combine TF-IDF using method described in Sec.~\ref{sec:decodingir}, obtaining significant performance improvement. 

\begin{table}[t]
\begin{center}
\begin{tabular}{|c|c|c|}
\hline
Models & R@1 & R@5 \\
\hline \hline
TF-IDF & 28.54 & 73.95 \\
\hline
AWI & 33.57  & 77.01 \\
\hline
AWI + TF-IDF & 40.70 & 85.39  \\
\hline
\end{tabular}
\end{center}
\caption{Retrieval results for the models using 1 in 10 recall rates (\%). 
\label{tab:ranking}} 
\end{table}

\section{Related work}
\label{sec:related}
Our work is related both to goal and non-goal oriented dialogue systems as the proposed model can be used as a language generation component in a goal-oriented dialogue \cite{Young2013POMDP} or simply to produce chit-chat style dialogue without a specific goal~\cite{Ritter2010Twitter,Banchs2012IRIS,Ameixa2014Movie}. 
Whereas traditionally a language generation component~\cite{Henderson2014RNN,Gasic2013Onlinepolicy,Wen2015EMNLP}  rely on explicit state~\cite{Williams2009SDS} in POMDP framework for goal-oriented dialogue system~\cite{Young2013POMDP}, the proposed model may relax such requirement. However, grounding the generated conversation with actions and knowledge is not studied in this paper. It will be a future work. 

The proposed model is related to the recent works in~\cite{Shang2015NRM,Vinyals2015NeuralConversationModel,Sordoni2015Conversation}, which use an encoder-decoder framework to model conversation. 
The closest work is in ~\cite{Sordoni2015Conversation}. This model differs from that work in using an attention model, an additional input similarity feature, and its decoder architecture. Importantly, this model is used not only for generation as in those previous work, but also for retrieval. 

Prior work to potentially increase specificity or diversity aims at producing multiple outputs~\cite{Carbonell1998MMR,Gimpel2013diversity} and our work is the same as in~\cite{Li2016MMI} to produce a single nontrivial output. Instead of using an objective function in \cite{Li2016MMI} that has an indirect relation to specificity, our model uses a specificity measure directly for training and decoding. 


\section{Conclusions}
We have presented a novel attentional neural conversation model with enhanced specificity using IDF. It has been evaluated for both response generation and retrieval. We have observed significant performance improvements in comparison to alternative methods. 

\bibliography{refs,cite-strings,cite-definitions}
\bibliographystyle{emnlp2016}

\end{document}